\documentstyle[11pt]{article}

\newcommand{\eclipse}{ECL$^i$PS$^e$}

\normalbaselines

\newcommand{\ES}{\mbox{$\emptyset$}}

\newcommand{\A}{\mbox{$\ \wedge\ $}}

\newcommand{\sse}{\mbox{$\:\subseteq\:$}}

\newcommand{\LL}{\mbox{$\ldots$}}

\newcommand{\C}[1]{\mbox{$\{{#1}\}$}}           

\newcommand{\NI}{\noindent}
\newcommand{\HB}{\hfill{$\Box$}}
\newcommand{\VV}{\vspace{5 mm}}
\newcommand{\III}{\vspace{3 mm}}
\newcommand{\II}{\vspace{2 mm}}

\newcommand{\szkew}[1]{\relax \setbox0=\hbox{\kern -24pt $\displaystyle#1$\kern 0pt }%
\box0}
{\catcode`\@=11 \global\let\ifjusthvtest@=\iffalse}

\newcounter{oldmycaption}

\newcommand{\p}[2]{\langle #1 \ ; \ #2 \rangle}

\newcommand{\ceiling}[1]{\lceil #1 \rceil}
\newcommand{\floor}[1]{\lfloor #1 \rfloor}

\newcounter{rulecnt}
\def\smallromani{\renewcommand{\theenumi}{\roman{enumi}}
\renewcommand{\labelenumi}{(\theenumi)}}

\newcommand{\Proof}{\NI
                    {\bf Proof.}\ }
\newtheorem{theorem}{Theorem}[section]
\newtheorem{defined}[theorem]{Definition}
\newenvironment{definition}{\begin{defined} \rm}{\end{defined}}
\newtheorem{exa}[theorem]{Example}
\newenvironment{example}{\begin{exa} \rm}{\end{exa}}

\newtheorem{claim}{Claim}
\newtheorem{exe}{Exercise}

\newtheorem{pro}{Problem}

\newcounter{symbol}
\newcommand{\indexsyma}[1]%
{\stepcounter{symbol}\index{zzz1 \thesymbol @\protect#1}}
\newcommand{\indexsymb}[1]%
{\stepcounter{symbol}\index{zzz2 \thesymbol @\protect#1}}
\newcommand{\indexsymc}[1]%
{\stepcounter{symbol}\index{zzz3 \thesymbol @\protect#1}}
\newcommand{\indexsymd}[1]%
{\stepcounter{symbol}\index{zzz4 \thesymbol @\protect#1}}
\newcommand{\indexsyme}[1]%
{\stepcounter{symbol}\index{zzz5 \thesymbol @\protect#1}}

\date{}

\title{A Proof Theoretic View of Constraint Programming}
\author{Krzysztof R. Apt
 \\
        {\it CWI } \\
        {\it P.O. Box 94079, 1090 GB Amsterdam, The Netherlands} \\
        and \\
        {\it  Dept. of Mathematics, Computer Science, Physics \& Astronomy } \\
        {\it University of Amsterdam, Plantage Muidergracht 24 } \\
        {\it 1018 TV Amsterdam, The Netherlands } \\
        {\it \verb+http://www.cwi.nl/~apt+}
}

\begin{document}
\maketitle

\begin{abstract} 
We provide here a proof theoretic account of constraint
programming that attempts to capture the essential ingredients of this
programming style.
We exemplify it by presenting proof rules for 
linear constraints over interval domains, and
illustrate their use by analyzing the constraint propagation process for
the {\tt SEND + MORE = MONEY} puzzle. 
We also show how this approach allows one to build new 
constraint solvers.

\end{abstract} 

\section{Introduction}

\subsection{Motivation}

One of the most interesting recent developments in the area of
programming has been constraint programming.  A prominent instance of it
is {\em constraint logic  programming} exemplified by such programming
languages as CLP(${\cal R}$), Prolog III or \eclipse{}. But recently
also imperative constraint programming languages emerged, such as 2LP of
\cite{MT95b} or CLAIRE of  \cite{CL96}. 
(For an overview of this area and related references see \cite{vHS96}).

The aim of this paper is to explain the essence of this approach to
programming without committing oneself to a particular programming
paradigm. We achieve this by providing a simple proof theoretic
framework that allows us in particular to explain  {\em constraint
propagation}, one of the cornerstones of  constraint programming.

The simplicity and elegance of constraint logic programming has already
led to a general presentation of their operational semantics in
\cite{JM94} that easily can be casted in a proof theoretic
jacket. But this account is limited to the logic programming view of
constraint programming. Moreover, it treats constraint propagation as a
further unexplained atomic action. 
Admittedly, the latter deficiency has been
addressed in \cite{Emd97}, where it has been explained how constraint
propagation can be defined within the framework of \cite{JM94}.

In our approach we try to ``decouple'' constraint programming from
logic programming by going back to the origins of constraints handling
and by viewing computing as a task of transforming one {\em constraint
  satisfaction problem\/} (CSP) into another, equivalent one. To take
into account reasoning by case analysis (that leads to ``don't-know''
nondeterminism and various forms of backtracking) we further introduce
a splitting operation that allows us to split one CSP into two, the
union of which is equivalent to the original CSP. The rules that
govern this process of transforming one CSP into a finite collection
of them fall naturally into four categories and seem to be sufficient
to describe the computation process.

By providing such a general view of constraint programming we can use
it to analyze the constraint programming process both within the logic
programming paradigm and the imperative one.  In particular, as we
shall see in Sections \ref{sec:lin} and \ref{sec:cs}, we can use it
both to study existing constraint solvers and to build new ones.
Additionally, as the Appendix shows, we can reason formally
about the proposed rules.

As a by-product of these considerations we bring constraint
programming closer to the {\em computation as deduction\/} paradigm
according to which the computation process is identified with a
constructive proof of a formula (a query) from a set of axioms. This
paradigm goes back to Herbrand and G\"{o}del and is exemplified by
logic programming and functional programming and also by viewing the
parsing process as a deduction (see e.g., \cite{SSP95}).

\subsection{Related Work}

A number of papers have advocated theorem proving as a means to
account for various aspects constraint logic programming. In
particular, in \cite{DG94} a Gentzen-style sequent calculus was used
to develop a logical semantics of constraint logic programs and 
in \cite{And95} proof-theoretic techniques were applied to
compare the intended theory and its actual implementation for various
constraint logic programming systems.

Next, 
a proof theoretic approach to constraint propagation
within the constraint logic programming framework has been proposed in
\cite{fruhwirth-constraint-95}. In this work so-called
constraint handling rules ({\sf CHR}) have been introduced. {\sf CHR} are
available as part of the \eclipse{} system and allow the user to
define his/her own constraint solvers.

Further, a related to ours approach to constraint programming has been
proposed in \cite{WKT95}. In this work constraint logic programs
are identified with so-called if-and-only-if definitions augmented
with integrity constraints. The if-and-only-if definitions are used to
define the ``usual'' logic programming computation step while the
integrity constraints are used to account for the constraint
propagation process that is identified with a complete resolution
strategy.
This approach is further elaborated and generalized in \cite{KTW98}.

Our view of constraint programming is also compatible with that
expounded in \cite{Smo96} where constraint programming
is presented without a commitment to a specific programming paradigm.
In fact, our approach allows one to couch his concepts of
propagators, inference engines and distributors
into a more specific, proof theoretic, framework.

The approach here presented
is closest to the one introduced independently in 
\cite{Cas98}. Even though the overall objectives are essentially
the same, the emphasis in his paper lies rather on defining  specific
techniques of constraint programming such as arc consistency and
forward checking by means of proof rules and strategies. 

Finally, let us mention the following unsubstantiated remark that
we found in  \cite[page 1115]{McA90}:
``In fact, virtually any form of constraint propagation can be defined
in terms of rules of inference''.
\subsection{Preliminaries}

We recall here the relevant definitions.
Consider a finite sequence of variables ${\cal X} := x_1, \LL, x_n$
where $n \geq 0$, with respective domains ${\cal D} := D_1, \LL, D_n$
associated with them.  So each variable $x_i$ ranges over the domain
$D_i$.  By a {\em constraint} $C$ on $\cal X$ we mean a subset of $D_1
\times \LL \times D_n$.  If $C$ equals $D_1 \times \LL \times D_n$
then we say that $C$ is {\em solved}. In the boundary case when the number 
$n$ of the variables equals 0
we admit two constraints, denoted by $\top$ and $\bot$, that
denote respectively the {\em true constraint\/} (for example $0 = 0$) and the
{\em false constraint\/} (for example $0 = 1$).

By a {\em constraint satisfaction problem}, {\em CSP\/} in short, 
we mean a triple
$\langle\cal X, \cal D, \cal C\rangle$, where $\cal C$ is a finite
set of constraints, each on a subsequence of $\cal X$.

Given a CSP $\langle\cal X, \cal D, \cal C\rangle$ with
${\cal X} := x_1, \LL, x_n$ and ${\cal D} := D_1, \LL, D_n$, 
we say that an $n$-tuple $(d_1, \LL, d_n) \in D_1 \times \LL \times D_n$
{\em is a solution to\/} $\langle\cal X, \cal D, \cal C\rangle$ if 
for every constraint $C \in {\cal C}$ on a sequence
$x_{i_1}, \LL ,x_{i_m}$ of the variables from ${\cal X}$ we have
\[
(d_{i_1}, \LL , d_{i_m}) \in C.
\]

Below we represent a CSP
$\langle\cal X, \cal D, \cal C\rangle$ as
an expression of the form
$\p{\cal C}{\cal DE}$, where 
${\cal DE} := \C{x_1 \in D_1, \LL, x_n \in D_n}$.
We call a construct of the form $x \in D$  a {\em domain expression}.
We stress the fact that a domain expression is not a constraint. 
By considering domain expressions separately, we can focus in the
sequel our attention on the proof rules that reduce domains.
Such rules are very common when dealing with linear constraints
and constraints on reals.

An alternative approach that we did not pursue here, is to dispense
with the domains by viewing each constraint as an $n$-ary relation and
by associating with each domain a unary constraint that coincides with
it. In this approach study of domain reduction becomes artificial.

To simplify the notation from now on we omit the ``\{ \}'' brackets
when presenting ${\cal C}$ and ${\cal DE}$.

We call a CSP {\em solved\/} if it is of the form $\p{\ES}{\cal DE}$
where no domain in ${\cal DE}$ is empty, and
{\em failed\/} if it either contains the false constraint $\bot$ or
some of its domains is empty. So a failed CSP admits no solution.

Given two CSP's $\phi$ and $\psi$, we call 
$\phi$ a {\em variant\/} of $\psi$ if the removal of solved constraints
from $\phi$ and $\psi$ yields the same CSP.

In what follows we assume that the constraints and the domain
expressions are defined in some specific, further unspecified,
language. In this representation of the constraints it is implicit
that each of them is a subset of the Cartesian product of the
associated variable domains.  For example, if we consider the CSP
$\p{x < y}{ x \in [0..10], y \in [5..10]}$, then we view the
constraint $x < y$ as the set $\C{(a,b) \mid a \in [0..10], b \in
  [5..10], a < b}$.

Given a constraint $c$ on the variables $x_1, \LL, x_n$
with respective domains $D_1, \LL, D_n$, and a sequence of 
domains $D'_1, \LL, D'_n$ such that 
for $i \in [1..n]$ we have $D'_i \sse D_i$, we say that
$c'$ is the {\em result of restricting $c$
to the domains $D'_1, \LL, D'_n$} if
$c' = c \cap (D'_1 \times \dots \times D'_n)$.

\section{The Proof Theoretic Framework}
In this section we introduce a proof theoretic framework that 
will be used throughout the paper.

\subsection{Format of the Proof Rules}
\label{subsec:pr}

In what follows we consider two types of proof rules that we call
{\em deterministic\/} and {\em splitting}.
The deterministic rules are of the form
\[
\frac{\phi}{\psi}
\]
where $\phi$ and $\psi$ are CSP's.
We assume here that $\phi$ is not failed and its set
of constraints is non-empty.
Depending on the form of the conclusion $\psi$
we distinguish two cases.
Assume that 
\[
\phi := \p{{\cal C}}{{\cal DE}}
\]
and
\[
\psi := \p{{\cal C'}}{{\cal DE'}}.
\]

\begin{itemize}

\item {\em Domain reduction rules}, or in short {\em reduction rules}.
These are rules in which 
the new domains are respective subsets of the old domains
and the new constraints  are respective restrictions 
of the old constraints to the new domains.

So here
${\cal DE} := x_1 \in D_1, \LL, x_n \in D_n$,
${\cal DE'} := x_1 \in D'_1, \LL, x_n \in D'_n$,
for $i \in [1..n]$ we have
$D'_i \sse D_i$, and  ${\cal C'}$ is the result 
of restricting each constraint in ${\cal C}$ to the
corresponding subsequence of the domains $D'_1, \LL,  D'_n$.

Here a failure is reached only when a domain of one or more variables gets
reduced to the empty set.

When all constraints in ${\cal C'}$ are solved, we call such a rule a
{\em solving rule}.

\item {\em Transformation rules}.
These rules are not domain reduction rules and
are such that
${\cal C'} \neq \ES$ and ${\cal DE} \sse {\cal DE'}$.  

The inclusion between ${\cal DE}$ and ${\cal DE'}$ means that the
domains of common variables are identical and that possibly new domain
expressions have been added to ${\cal DE}$.  Such new domain
expressions deal with new variables on which some constraints have
been introduced.  

Here a failure is reached only when the false constraint is generated.

\end{itemize}

The splitting rules are of the form
\[
\frac{\phi}{\psi_1 \mid \psi_2}
\]
where  $\phi, \psi_1$ and $ \psi_2$ are CSP's.
As for deterministic
rules we assume here that $\phi$ is not failed and its set of
constraints is non-empty.
In what follows we only consider splitting rules in which
$\phi, \psi_1$ and $ \psi_2$ are CSP's
with the same sequence of variables.

These rules allow us to replace one
CSP by two CSP's. The intuition is that
 their ``union'' is ``equivalent'' to the original CSP.
They are counterparts of the rules just introduced. So, again,
we distinguish two cases.

\begin{itemize}

\item {\em Reduction splitting rules}. These are rules such that
both $\frac{\psi}{\phi_1}$ and $\frac{\psi}{\phi_2}$ are
reduction rules.

\item {\em Transformation splitting rules}. These are rules such that
both $\frac{\psi}{\phi_1}$ and $\frac{\psi}{\phi_2}$ are
transformation rules.

\end{itemize}

\subsection{Examples of Proof Rules}
\label{subsec:exapr}

In the sequel when presenting specific proof rules we
delete from the conclusion all solved constraints.
Also, we abbreviate the domain expression $x \in \C{a}$ to $x = a$.

As an example of a reduction rule consider the following rule:
\begin{list}%
{}{\usecounter{rulecnt}}

\centering
\item {\em EQUALITY \ref{eq:eq0}}
\label{eq:eq0}

\[
\frac
{\p{x = y}{x \in D_1, y \in D_2}}
{\p{x = y}{x \in D_1 \cap D_2, y \in D_1 \cap D_2}}
\]
\end{list}

Note that this rule yields a failure when $D_1 \cap D_2 = \ES$.
In case  $D_1 \cap D_2$ is a singleton this rule becomes a solving rule,
that is the constraint $x=y$ becomes solved (and hence deleted).
Note also the following solving rule:

\begin{list}%
{}{\usecounter{rulecnt}}
\setcounter{rulecnt}{1}
\centering
\item {\em EQUALITY \ref{eq:eq1}}
\label{eq:eq1}
\[
\frac
{\p{x = x}{x \in D}}
{\p{}{x \in D}}
\]
\end{list}

Following the just introduced convention we dropped the constraint
from the conclusion of the {\em EQUALITY \ref{eq:eq1}}
rule. This explains its format.  

As further examples of solving rules
consider the following three concerning disequality:

\begin{list}%
{}{\usecounter{rulecnt}}

\centering
\item {\em DISEQUALITY \ref{eq:diseq1}}
\label{eq:diseq1}
\[
\frac
{\p{x \neq x}{x \in D}}
{\p{}{x \in \ES}}
\]

\item {\em DISEQUALITY \ref{eq:diseq}}
\label{eq:diseq}
\[
\frac
{\p{x \neq y}{x \in D_1, y \in D_2}}
{\p{}{x \in D_1, y \in D_2}}
\]  

\end{list}
where $D_1 \cap D_2 = \ES$,

\begin{list}%
{}{\usecounter{rulecnt}}
\setcounter{rulecnt}{2}

\centering

\item {\em  DISEQUALITY  \ref{rule:diseq4}}
\label{rule:diseq4}
\[
\frac
{\p{x \neq y}{x \in D, y = a}}
{\p{}{x \in D - \C{a}, y = a}}
\]
\end{list}
where $a \in D$, and similarly with $x \neq y$ replaced by $y \neq x$.

So the {\em DISEQUALITY \ref{eq:diseq1}} rule yields a failure
while the {\em DISEQUALITY \ref{rule:diseq4}} rule can yield a failure.

Next, as an example of a transformation rule consider the
following rule that substitutes a variable by a value:

\begin{list}%
{}{\usecounter{rulecnt}}
\centering
\item {\em SUBSTITUTION}  
  \label{eq:subst}
\[
\frac
{\p{{\cal C}}{{\cal DE}, x = a}}
{\p{{\cal C} \C{x/ \overline{a}}}{{\cal DE}, x = a}}
\]
\end{list}
where $x$ occurs in ${\cal C}$.

Here $\overline{a}$ stands for the constant that denotes
in the underlying language the value $a$ and
${\cal C} \C{x/ \overline{a}}$ denotes the set of constraints obtained from
${\cal C}$ by substituting in it every occurrence of $x$ by $\overline{a}$.
So $x$ does not occur in ${\cal C} \C{x/ \overline{a}}$.

Another example of a transformation rule forms the following rule:
\begin{list}%
{}{\usecounter{rulecnt}}
\centering
\item {\em DELETION}

\[
\frac
{\p{{\cal C} \cup \C{\top}}{{\cal DE}}}
{\p{{\cal C}}{{\cal DE}}}
\]
\end{list}

Let us consider now the splitting rules. A natural class of examples
of reduction splitting rules form
{\em domain splitting rules}.
These are rules of the form:

\[
\frac
{\p{{\cal C}}{{\cal DE}, x \in D}}
{\p{{\cal C'}}{{\cal DE}, x \in D_1} \mid 
 \p{{\cal C''}}{{\cal DE}, x \in D_2}}
\]
where $D_1 \cup D_2 = D$, $D_i \neq \ES$ for $i \in \C{1,2}$, 
${\cal C'}$ is the result 
of restricting each constraint in ${\cal C}$ to the
corresponding subsequence of the 
domains in ${\cal DE}$ and $D_1$, and analogously with ${\cal C''}$.

If such a rule does not depend on ${\cal C}$ and ${\cal DE}$ 
we abbreviate it to
\[
\frac
{x \in D}
{x \in D_1 \mid  x \in D_2}
\]
Two specific instances are:

\begin{list}%
{}{\usecounter{rulecnt}}
\centering
\item {\em ENUMERATION}
\[
\frac{x \in D}{x = a \mid x \in D - \C{a}}
\]
\end{list}
where $D$ is a finite domain with at least two elements and $a \in D$, and

\begin{list}%
{}{\usecounter{rulecnt}}
\centering
\item {\em BISECTION}
\[
\frac{x \in [a..b]}{x \in [a .. \frac{a+b}{2}] \mid x \in [\frac{a+b}{2} .. b]}
\]
\end{list}
where $[a..b]$ a closed non-empty interval of reals.
Here we wish to preserve the property that the intervals are closed so 
the new intervals are not disjoint.

Finally, a natural class of examples of transformation splitting rules
form {\em constraint splitting rules}.
They have the following form:

\[
\frac
{\p{{\cal C}, C}{{\cal DE}}}
{
\p{{\cal C}, C_1}{{\cal DE}} \mid \p{{\cal C}, C_2}{\cal DE}}
\]
where
\begin{itemize}

\item every solution to ${\p{{\cal C}, C}{{\cal DE}}}$
is a solution to 
$\p{{\cal C}, C_1}{{\cal DE}}$ or $\p{{\cal C}, C_2}{{\cal DE}}$,
\item every solution to $\p{{\cal C}, C_i}{{\cal DE}}$
($i \in [1,2]$) 
is a solution to ${\p{{\cal C}, C}{{\cal DE}}}$.
\end{itemize}

If such a rule does not depend on ${\cal C}$ and ${\cal DE}$
we abbreviate it to

\[
\frac{
C
}{
C_1 \mid C_2
}
\]

A particular instance is:
\[
\frac{|x-y|=a}{x -y = a \mid x - y = -a}
\]
where $x$ and $y$ are integer variables and $a$ is an integer.

\subsection{Derivations}

Now that we have defined the proof rules, we
define the result
of applying a proof rule to a CSP. Assume
a CSP of the form $\p{{\cal C}  \cup {\cal C}_1}{{\cal D} \cup {\cal D}_1}$.
First, consider a deterministic rule, so a rule of the form
\begin{equation}
\frac
{\p{{\cal C}_1}{{\cal D}_1}}
{\p{{\cal C}_2}{{\cal D}_2}}
\label{eq:det}
\end{equation}

Here a clash of variables can take place if some variable
of ${\cal C}_2$ also appears in ${\cal C}$ but not in ${\cal C}_1$.
Then such a variable of ${\cal C}_2$ should be renamed first.
So let us rename the variables of 
$\p{{\cal C}_2}{{\cal D}_2}$
that appear in ${\cal C}$ but not in ${\cal C}_1$
by some fresh variables 
and denote the so obtained CSP by
$\p{{\cal C}^{'}_2}{{\cal D}^{'}_2}$.

We say that rule (\ref{eq:det}) {\em can be applied\/} to
$\p{{\cal C}  \cup {\cal C}_1}{{\cal D} \cup {\cal D}_1}$
and call
\[
\p{{\cal C}  \cup {\cal C}^{'}_2}{{\cal D} \cup {\cal D}^{'}_2}
\]
the {\em result of applying rule} (\ref{eq:det}) {\em to} $\p{{\cal C}
  \cup {\cal C}_1}{{\cal D} \cup {\cal D}_1}$.  If $\p{{\cal C} \cup
  {\cal C}^{'}_2}{{\cal D} \cup {\cal D}^{'}_2}$ is not a variant of
$\p{{\cal C} \cup {\cal C}_1}{{\cal D} \cup {\cal D}_1}$, then we say
that it is the result of a {\em relevant application of rule}
(\ref{eq:det}) {\em to} $\p{{\cal C} \cup {\cal C}_1}{{\cal D} \cup
{\cal D}_1}$.

Further, given a CSP $\phi$ and a deterministic rule $R$, we say
that $\phi$ is {\em closed under the applications of $R$\/} if
either $R$ cannot be applied to $\phi$ or no application of it to
$\phi$ is relevant.

For example, assume for a moment the expected interpretation of 
propositional formulas and consider the CSP 
$\phi := \p{x \A y = z}{x = 1, y = 0, z = 0}$.
Here $x=1$ is an abbreviation for the domain expression $x \in \C{1}$ and
similarly for the other variables.

This CSP
is closed under the applications of the transformation rule
\[
\frac
{\p{x \A y = z}{x = 1, y \in D_y, z \in D_z}}
{\p{z = y}{x = 1, y \in D_y, z \in D_z}}
\]
Indeed, this rule can be applied to $\phi$; the outcome is 
$\psi := \p{z = y}{x = 1, y = 0, z = 0}$. After the removal 
of solved constraints from $\phi$ and $\psi$ we get in both cases the
solved CSP 
$\p{\ES}{x = 1, y = 0, z = 0}$.

In contrast, the CSP
$\phi := \p{x \A y = z}{x = 1, y \in \C{0,1}, z \in \C{0,1}}$
is not closed under the applications of the above rule because 
$\p{z = y}{x = 1, y \in \C{0,1}, z \in \C{0,1}}$ is not a variant of $\phi$.

Next, consider a splitting rule, so
a rule of the form

\begin{equation}
\frac
{\p{{\cal C}_1}{{\cal D}_1}}
{\p{{\cal C}_2}{{\cal D}_2} \mid 
 \p{{\cal C}_3}{{\cal D}_3}}
\label{eq:split}
\end{equation}

We then say that rule (\ref{eq:split}) {\em can be applied\/} to
$\p{{\cal C}  \cup {\cal C}_1}{{\cal D} \cup {\cal D}_1}$
and call
\begin{equation}
\p{{\cal C}  \cup {\cal C}_2}{{\cal D} \cup {\cal D}_2} \mid
\p{{\cal C}  \cup {\cal C}_3}{{\cal D} \cup {\cal D}_3} 
\label{eq:res}
\end{equation}
the {\em result of applying it to}
$\p{{\cal C}  \cup {\cal C}_1}{{\cal D} \cup {\cal D}_1}$.
If neither
$\p{{\cal C}  \cup {\cal C}_2}{{\cal D} \cup {\cal D}_2}$
nor 
$\p{{\cal C}  \cup {\cal C}_3}{{\cal D} \cup {\cal D}_3}$
is a variant of
$\p{{\cal C} \cup {\cal C}_1}{{\cal D} \cup {\cal D}_1}$, then we say
that  (\ref{eq:res})
 is the result of a {\em relevant application of rule} 
  (\ref{eq:split}) {\em to} $\p{{\cal C} \cup {\cal C}_1}{{\cal D} \cup
{\cal D}_1}$.
(Recall that by assumption all three CSP's ${\p{{\cal C}_i}{{\cal D}_i}}$, 
where $i \in [1..3]$, have the same sequence of variables,
so we do not need to worry here about variable clashes.)

Finally, we introduce the notions of a proof tree and of a derivation.

\begin{definition}
Assume a set of proof rules.
A {\em proof tree\/} is a tree the nodes of which are CSP's.
Further, each node has at most two direct descendants and 
for each node $\phi$ the following holds:
    \begin{itemize}

    \item If $\phi$ is a leaf, then no application of
      a rule to $\psi$ is relevant;
    \item If $\phi$ has precisely one direct descendant, say $\psi$, then
      $\psi$ is the result of a relevant application of a proof rule
      to $\phi$;
    \item If $\phi$ has precisely two direct descendants, say $\psi_1$
      and $\psi_2$, then $\psi_1 \mid \psi_2$ is the result of a
      relevant application of a proof rule to $\phi$.
    \end{itemize}

    A {\em derivation\/} is a branch in a proof tree.  A derivation is
    called {\em successful\/} if it is finite and its last element is
    a solved CSP. A derivation is called {\em failed\/} if it is
    finite and its last element is a failed CSP.  
\HB
\end{definition}

The idea behind the above definition is 
 that we consider in the proof trees only 
those applications of the proof rules that cause some change.
Note also that more proof rules can be applicable to a given CSP, so
a specific CSP can be a root of several proof trees.

Note that some finite derivations are neither successful nor failed.
In fact, many constraint solvers yield CSP's that are neither solved
nor failed --- their aim is to bring the initial CSP to some specific,
simpler form.

In some cases this third possibility does not arise. Indeed, consider a
non-failed CSP with a non-empty set of constraints on finite domains.
Then either
the {\em DELETION\/} or the {\em SUBSTITUTION\/} or the {\em
ENUMERATION\/} rule can be applied to it and moreover each such application
is always relevant. So in presence of the above three rules for CSP's
with finite domains each finite derivation is either successful or
failed.

\subsection{Equivalent CSP's}

We introduced the proof rules so that we can
reduce one CSP to another CSP or to two CSP's, which are 
in some sense ``smaller'' yet ``equivalent''. 
Both notions can be made precise but the first one will not play
any role in our considerations, so we only present an adequate notion of 
equivalence.  Because the considered proof rules are of a specific
form, we limit ourselves in the definition to specific pairs of CSP's.

\begin{definition}
Consider two CSP's $\phi$ and $\psi$ such that
all variables of $\phi$ are also present in $\psi$. 
We say that the CSP's $\phi$ and $\psi$ are 
{\em equivalent \/} if
\begin{itemize}
\item every solution to $\phi$ is or can be extended to a solution to $\psi$,
\item for every solution to $\psi$ its restriction to the variables of  $\phi$
 is a solution to $\phi$.
\HB
\end{itemize}
\end{definition}

In particular, two CSP's with the same sequence of variables are
equivalent if they have the same set of solutions.
So for example the CSP's 
\[
{\p{3x - 5y = 4}{x \in [0..9], y \in [1..8]}}
\]
and
\[
{\p{3x - 5y = 4}{x \in [3..8], y \in [1..4]}}
\]
are equivalent, since both of them have 
$x=3, y=1$ and $x=8, y=4$ as the only solutions,
and, for ${\cal DE} := x \in D_x, y \in D_y, z \in D_z$,
so are
\[
{\p{x<y, y<z}{{\cal DE}}}
\]
and
\[
{\p{x<y, y<z, x<z}{{\cal DE}}}.
\]

In contrast,
\[
{\p{x<z}{x \in D_x, z \in D_z}}
\]
and
\[
{\p{x<y, y<z}{{\cal DE}}}
\]
are not equivalent, as not each solution to the former 
extends to a solution of the latter.

This brings us to the
following notion where we make use of the fact that the considered
proof rules are of a specific form.

\begin{definition} \mbox{} \\
\vspace{-6mm}
  
  \begin{enumerate} \smallromani
  \item 
A proof rule 
\[
\frac{\phi}{\psi}
\]
is called 
{\em equivalence preserving\/} if $\phi$ and $\psi$ 
are equivalent.

\item 
A proof rule
\[
\frac{\phi}{\psi_1 \mid \psi_2}
\]
is called 
{\em equivalence preserving\/} if 

\begin{itemize}
\item every solution to $\phi$ is a solution to $\psi_1$ or to $\psi_2$,
\item every solution to $\psi_i$ ($i \in [1,2]$) is a solution to  $\phi$.
\HB

\end{itemize}
  \end{enumerate}
\end{definition}

All the rules discussed so far are
equivalence preserving.  From the way we introduce the proof rules
in the sequel it will be clear that all of them are also 
equivalence preserving.

This completes the presentation of our proof theoretic framework.  In
order to use it to model constraint programming the proof rules above
introduced have to be ``customized'' to a specific language in which
constraints are defined and to specific domains.  In what follows we
present an example of such a customization that deals with linear
constraints over interval and finite domains.

Such rules should be selected and scheduled in an
appropriate way and some strategy should be employed to traverse the
generated proof trees. We defer discussion of these issues to
Subsection \ref{subsec:control}.

\section{Linear Constraints over Interval Domains}
\label{sec:lin}

In this section we consider linear constraints over interval domains.
We use the introduced rules to discuss the behaviour of the \eclipse{}
finite domain solver and, by means of an example, 
to analyse the {\em SEND + MORE = MONEY} puzzle.

First, let us recall the relevant definitions.
By a {\em linear expression\/} we mean a term in the language that
contains two constants 0 and 1, the unary minus function $-$ and two binary
functions + and $-$, both written in the infix notation. We abbreviate
terms of the form 
\[
\underbrace{1 +  \LL + 1}_{\mbox{$n$ times}}
\]
to $n$, terms of the form
\[
\underbrace{x +  \LL + x}_{\mbox{$n$ times}}
\]
to $n x$ and analogously with $-1$ and $-x$ used instead of $1$ and $x$.
So (using appropriate transformation rules)
each linear expression can be equivalently written in the form 
\[
a_1 x_1 + \LL + a_n x_n + a_{n+1}  
\]
where $n \geq 0$, $a_1, \LL ,a_n$ are non-zero integers,
$x_1, \LL ,x_n$ are different variables and $a_{n+1}$ is an integer.

By a {\em linear constraint\/} we mean a formula of the
form
\[
s \: op \: t
\]
where $s$ and $t$ are linear expressions and $op \in \C{<, \leq, =, \neq, \geq, >}$.
In what follows we drop the qualification
``linear'' when discussing linear expressions and linear constraints.

Further, we call
\begin{itemize}

\item $s < t $ and $s > t$ {\em strict inequality constraints},

\item $s \leq t $ and $s \geq t$ {\em  inequality constraints},

\item $s = t$ an {\em  equality constraint}, 

\item $s \neq t$ a {\em  disequality constraint},

\item $x \neq y$, for variables $x,y$,
a {\em simple  disequality constraint}.
\end{itemize}

By an {\em integer interval}, or an {\em interval\/} in short, we mean
an expression of the form
\[
[a..b]
\]
where $a$ and $b$ are integers;
$[a..b]$ denotes the set of all integers between $a$ and $b$,
including $a$ and $b$.
If $a > b$, we call $[a..b]$ the {\em empty interval}.

Finally, by a {\em range\/} we mean an expression of the form
\[
x \in I
\]
where $x$ is a variable and $I$ is an interval.
We abbreviate $x \in [a..b]$ to $x=a$ 
if $a=b$ and write $x \in \ES$ if $a > b$.

In what follows we  discuss various rules that allow us to manipulate
linear constraints over interval domains.
We assume that all considered linear constraints have at least one variable.

\subsection{Reduction Rules for  Inequality Constraints}
\label{subsec:inequ}

We begin with the inequality constraints.
Using appropriate transformation rules
each  inequality constraint can be equivalently written in the form
\begin{equation}
\sum_{i \in {\em POS}} a_i x_i -  \sum_{i \in {\em NEG}} a_i x_i \leq b
  \label{eq:n-lin}
\end{equation}
where 
\begin{itemize}
\item $a_i$ is a positive integer for $i \in {\em POS} \cup {\em NEG}$,

\item $x_i$ and $x_j$ are different variables for $i \neq j$ and $i,j
  \in {\em POS} \cup {\em NEG}$,

\item $b$ is an integer.
\end{itemize}

Assume the ranges
\[
x_i \in [l_i .. h_i]
\]
for $i \in {\em POS} \cup {\em NEG}$.

Choose now some $j \in POS$ and let us rewrite (\ref{eq:n-lin}) as 
\[
x_j  \leq \frac{b - \sum_{i \in {\em POS} - \{j\}} a_i x_i +  \sum_{i \in {\em NEG}} a_i x_i}{a_j}
\]
Computing the maximum of the 
expression on the right-hand side w.r.t. the ranges of the involved variables
we get
\[
x_j  \leq \alpha_j
\]
where
\[
\alpha_j := \frac{b - \sum_{i \in {\em POS} - \{j\}} a_i l_i +  \sum_{i \in {\em NEG}} a_i h_i}{a_j}
\]
so, since the variables assume integer values,
\[
x_j  \leq \floor{\alpha_j}.
\]
We conclude that
\[
x_j  \in [l_j .. min(h_j, \floor{\alpha_j})].
\]

By analogous calculations we conclude for $j \in {\em NEG}$ 
\[
x_j  \geq \ceiling{\beta_j}
\]
where
\[
\beta_j := \frac{- b + \sum_{i \in {\em POS}} a_i l_i -  \sum_{i \in {\em NEG} - \{j\}} a_i h_i}{a_j}
\]
In this case we conclude that
\[
x_j  \in [max(l_j, \ceiling{\beta_j}) .. h_j].
\]

This brings us to the following reduction rule for inequality constraints:

\begin{list}%
{}{\usecounter{rulecnt}}
\centering
\item {\em LINEAR INEQUALITY \ref{eq:lin}}
\label{eq:lin}

\[
\frac
{\p{\sum_{i \in {\em POS}} a_i x_i -  \sum_{i \in {\em NEG}} a_i x_i \leq b}{x_1 \in [l_1 .. h_1],  \LL,  x_{n} \in [l_n..h_n]}}
{\p{\sum_{i \in {\em POS}} a_i x_i -  \sum_{i \in {\em NEG}} a_i x_i \leq b}{x_1 \in [l'_1 .. h'_1],  \LL,  x_{n} \in [l'_n..h'_n]}}
\]
\end{list}
where for $j \in {\em POS}$
\[
l'_j := l_j, \ h'_j :=  min(h_j, \floor{\alpha_j})
\]
and for $j \in {\em NEG}$
\[
l'_j := max(l_j, \ceiling{\beta_j}),  \ h'_j := h_j.
\]

\subsection{Reduction Rules for  Equality Constraints}
\label{subsec:equ}

Each equality constraint can be equivalently written
as two inequality constraints.
By combining the corresponding reduction rules for these two  inequality
constraints we obtain a reduction rule for an equality constraint.
More specifically, each equality constraint can be equivalently written in the form

\begin{equation}
  \label{eq:equality}
\sum_{i \in {\em POS}} a_i x_i -  \sum_{i \in {\em NEG}} a_i x_i = b
\end{equation}
where we adopt the conditions that follow (\ref{eq:n-lin})
in the previous subsection.

Assume now the ranges
\[
{x_1 \in [l_1 .. h_1],  \LL,  x_{n} \in [l_n..h_n]}.
\]
We infer then both the conclusion of the {\em LINEAR INEQUALITY \ref{eq:lin}} 
reduction rule and 
\[
{x_1 \in [l''_1 .. h''_1],  \LL,  x_{n} \in [l''_n..h''_n]}
\]
where for $j \in {\em POS}$
\[
l''_j := max(l_j, \ceiling{\gamma_j}), \ h''_j := h_j
\]
with 
\[
\gamma_j := \frac{b - \sum_{i \in {\em POS} - \{j\}} a_i h_i +  \sum_{i \in {\em NEG}} a_i l_i}{a_j}
\]
and for $j \in {\em NEG}$
\[
l''_j := l_j,  \ h''_j := min(h_j, \floor{\delta_j})
\]
with
\[
\delta_j := \frac{- b + \sum_{i \in {\em POS}} a_i h_i -  \sum_{i \in {\em NEG} - \{j\}} a_i l_i}{a_j}
\]

This yields the following reduction rule:

\begin{list}%
{}{\usecounter{rulecnt}}
\centering
\item {\em LINEAR EQUALITY }
\label{eq:equ}
\[
\frac
{\p{\sum_{i \in {\em POS}} a_i x_i -  \sum_{i \in {\em NEG}} a_i x_i = b}{x_1 \in [l_1 .. h_1],  \LL,  x_{n} \in [l_n..h_n]}}
{\p{\sum_{i \in {\em POS}} a_i x_i -  \sum_{i \in {\em NEG}} a_i x_i = b}{x_1 \in [l'_1 .. h'_1],  \LL,  x_{n} \in [l'_n..h'_n]}}
\]
\end{list}
where for $j \in {\em POS}$
\[
l'_j := max(l_j, \ceiling{\gamma_j}), \ h'_j :=  min(h_j, \floor{\alpha_j})
\]
and for $j \in {\em NEG}$
\[
l'_j := max(l_j, \ceiling{\beta_j}),  \ h'_j := min(h_j, \floor{\delta_j}).
\]

As an example of the use of the above reduction rule
consider the CSP
$\p{3x - 5y = 4}{x \in [0..9], y \in [1..8]}$.
A straightforward calculation shows that 
$x \in [3..9], \ y \in [1..4]$
are the ranges in the conclusion of {\em LINEAR EQUALITY } rule.
Another application of the rule yields the ranges
$x \in [3..8]$ and $y \in [1..4]$
upon which the process stabilizes.

Note that if in (\ref{eq:equality}) there is only one variable,
the {\em LINEAR EQUALITY } rule reduces to the following solving rule:
\[
\frac
{\p{ax = b}{x \in [l..h]}}
{\p{}{x \in \C{\frac{b}{a}} \cap [l..h]}}
\]
So, if $a$ divides $b$ and $l \leq \frac{b}{a} \leq h$, the
domain expression  $x = \frac{b}{a}$ is inferred,
and otherwise a failure is reached.

\subsection{Transformation Rules for Inequality and Equality Con\-straints}

The above reduction rules can be applied only to
inequality and equality constraints
that are in a specific form, (\ref{eq:n-lin}) or (\ref{eq:equality}).
So we need to augment the introduced reduction rules by appropriate transformation rules.
Depending on the level
of description one can content oneself with a couple of general rules
or several very detailed ones.
These rules are pretty straightforward and are omitted.

\subsection{Rules for Disequality Constraints}
\label{subsec:diseq}

The reduction rules for simple disequalities are very natural.
First, note that the following rule

\begin{list}%
{}{\usecounter{rulecnt}}
\centering
\item {\em  SIMPLE DISEQUALITY  \ref{rule:diseq1}}
\label{rule:diseq1}
\[
\frac
{\p{x \neq y}{x \in [a..b], y \in [c..d]}}
{\p{}{x \in [a..b], y \in [c..d]}}
\]
\end{list}
where $b < c$ or $d < a$, is an instance of the
{\em DISEQUALITY \ref{eq:diseq}}  solving rule
introduced in Subsection \ref{subsec:exapr}
and where following the convention there mentioned
we dropped the constraint and the
domain expressions from the conclusion of the proof rule.

Next, we adopt the following two solving rules that
are instances of the solving {\em DISEQUALITY \ref{rule:diseq4}} rule:

\begin{list}%
{}{\usecounter{rulecnt}}
\setcounter{rulecnt}{1}
\centering

\item {\em  SIMPLE DISEQUALITY  \ref{rule:diseq2}}
\label{rule:diseq2}
\[
\frac
{\p{x \neq y}{x \in [a..b], y = a}}
{\p{}{x \in [a+1 .. b], y = a}}
\]

\item {\em  SIMPLE DISEQUALITY  \ref{rule:diseq3}}
\label{rule:diseq3}
\[
\frac
{\p{x \neq y}{x \in [a..b], y = b}}
{\p{}{x \in [a .. b-1], y = b}}
\]
\end{list}
and similarly with $x \neq y$ replaced by $y \neq x$.
Recall that the domain expression $y = a$ is a shorthand for $y \in [a..a]$.

To deal with disequality constraints that are not simple ones
we use the following notation.
Given a linear expression $s$ and a sequence of ranges involving all
the variables of $s$ we denote by $s^{-}$ the minimum $s$ can take
w.r.t. these ranges and by $s^{+}$ the maximum $s$ can take
w.r.t. these ranges. The considerations of Subsection \ref{subsec:inequ}
show how $s^{-}$ and $s^{+}$ can be computed.

We now introduce the following transformation rule for 
non-simple disequality constraints:

\begin{list}%
{}{\usecounter{rulecnt}}
\setcounter{rulecnt}{2}
\centering
\item {\em DISEQUALITY \ref{eq:dis-ns}}
\label{eq:dis-ns}
\[
\frac
{\p{s \neq t}{{\cal DE}}}
{\p{x \neq t, \ x = s}{x \in [s^{-} .. s^{+}],{\cal DE}}}
\]
\end{list}
where 
\begin{itemize}

\item $s$ is not a variable,

\item $x$ is a fresh variable,

\item ${\cal DE}$ is a sequence of the ranges involving the variables present in $s$ and $t$,

\item  $s^{-}$ and $s^{+}$ are computed w.r.t. the ranges in ${\cal DE}$.
\end{itemize}

An analogous rule is introduced for the inequality $s \neq t$, where
$t$ is not a variable.

\subsection{Domain splitting rules}

We conclude our treatment of linear constraints over interval domains
by presenting three specific domain splitting rules. Because at
the moment the
assumed domains are intervals, these rules
so designed that the domains remain intervals.

\begin{list}%
{}{\usecounter{rulecnt}}
\setcounter{rulecnt}{0}
\centering
\item {\em INTERVAL SPLITTING \ref{eq:int1}}
\label{eq:int1}
\[
\frac
{x \in [a..b]}
{x = a \mid x \in [a+1 .. b]}
\]
\end{list}
where $a < b$,

\begin{list}%
{}{\usecounter{rulecnt}}
\setcounter{rulecnt}{1}
\centering
\item {\em INTERVAL SPLITTING \ref{eq:int2}}
\label{eq:int2}
\[
\frac
{x \in [a..b]}
{x = b \mid x \in [a .. b-1]}
\]
\end{list}
where $a<b$,

\begin{list}%
{}{\usecounter{rulecnt}}
\setcounter{rulecnt}{2}
\centering
\item {\em INTERVAL SPLITTING \ref{eq:int3}}
\label{eq:int3}
\[
\frac
{\p{x \neq c}{x \in [a..b]}}
{\p{}{x \in [a..c-1]} \mid \p{}{x \in [c+1..b]}}
\]
\end{list}
where $a < c < b$.

Finally, to deal with the strict inequality constraints it suffices to
use expected transformation rules that reduce them to inequalities
and disequalities.

\subsection{Shifting from Intervals to Finite Domains}

In our presentation we took care that all the rules preserved the
property that the domains are intervals.
In some systems, such as \eclipse{}, this property is relaxed and
instead of finite intervals finite sets of integers are chosen. 
To model the use of such finite domains
it suffices to modify some of the rules introduced above.

In the case of inequality constraints
we can use the following minor modification of the 
{\em LINEAR INEQUALITY \ref{eq:lin}} reduction rule:

\begin{list}%
{}{\usecounter{rulecnt}}
\setcounter{rulecnt}{1}
\centering
\item {\em LINEAR INEQUALITY \ref{eq:lin2}}
\label{eq:lin2}
\[
\frac
{\p{\sum_{i \in {\em POS}} a_i x_i -  \sum_{i \in {\em NEG}} a_i x_i \leq b}{x_1 \in D_1, \LL, x_n \in D_n}}
{\p{\sum_{i \in {\em POS}} a_i x_i -  \sum_{i \in {\em NEG}} a_i x_i \leq b}{x_1 \in [l'_1 .. h'_1] \cap D_1,  \LL,  x_{n} \in [l'_n..h'_n] \cap D_n}}
\]
\end{list}
where $l'_j$ and $h'_j$ are defined as in  the 
{\em LINEAR INEQUALITY \ref{eq:lin}} reduction
rule  with $l_j := min(D_j)$ and $h_j :=  max(D_j)$.

Note that in this rule the domains are now arbitrary finite sets
of integers.
An analogous modification can be introduced for the case of
the reduction rule for equality constraints.

In the case of a simple disequality constraint we use the
{\em  DISEQUALITY  \ref{rule:diseq4}} solving rule.
So now, in contrast to the case of interval domains, an arbitrary
element can be removed from a domain, not only the ``boundary'' one.

\subsection{Example: the {\em SEND + MORE = MONEY} Puzzle}

We now illustrate use of the above rules 
for linear constraints over interval and finite domains by analyzing in detail
the well-known {\em SEND + MORE = MONEY}  puzzle.
Recall that this puzzle calls for a solution of the equality constraint
\II

\[
\begin{array}{ll}
  & 1000 \cdot S + 100 \cdot E + 10 \cdot N + D \\
+ & 1000 \cdot M + 100 \cdot O + 10 \cdot R + E \\
= & 10000 \cdot M + 1000 \cdot O + 100 \cdot N + 10 \cdot E + Y
\end{array}
\]

\NI
together with
28 simple disequality constraints $x \neq y$ for $x, y \in \C{S,E,N,D,M,O,R,Y}$
where $x$ preceeds $y$ in the alphabetic order,
and with the range $[1..9]$ for $S$ and $M$ and the range $[0..9]$ for
the other variables.

Both in the CHIP system (see \cite[page 143]{VANHENTENRYCK89A}) and
in \eclipse{} Version 3.5.2. the above CSP is internally reduced to
the one with the following domain expressions:
\begin{equation}
  \label{eq:send}
S = 9, E \in [4..7], N \in [5..8],  D \in [2..8],  M = 1,  O = 0,  R \in [2..8],  Y \in [2..8].  
\end{equation}
We now show how this outcome can be
formally derived using the rules we introduced.

First, using the transformation rules for linear constraints we can 
transform the above equality to
\[
9000 \cdot M + 900 \cdot O + 90 \cdot N + Y - (91 \cdot E + D + 1000 \cdot S + 10 \cdot R) = 0.
\]
Applying the {\em LINEAR EQUALITY \/}  reduction rule 
with the initial ranges we obtain the following sequence of new
ranges:
\[
S = 9, E \in [0..9], N \in [0..9],  D \in [0..9],  M = 1,  O \in [0..1],  R \in [0..9],  Y \in [0..9].
\]

At this stage a subsequent use of the same rule yields no new outcome.
However, by virtue of the fact that $M = 1$ we can now apply the {\em
  SIMPLE DISEQUALITY \ref{rule:diseq3}\/} solving rule to
$M \neq O$ to conclude that $O = 0$.  Using now
the facts that $M=1, O = 0, S = 9$, the solving rules 
{\em  SIMPLE DISEQUALITY  \ref{rule:diseq2}} and {\em \ref{rule:diseq3}}
can be repeatedly applied to shrink the ranges of the
other variables.  This yields the following new sequence of ranges:
\[
S = 9, E \in [2..8], N \in [2..8],  D \in [2..8],  M = 1,  O = 0,  R \in [2..8],  Y \in [2..8].
\]

Now five successive iterations of the {\em LINEAR EQUALITY \/} 
reduction rule 
yield the following sequences of shrinking ranges of $E$ and $N$ with other ranges
unchanged:

\[
E \in [2..7], N \in [3..8],
\]

\[
E \in [3..7], N \in [3..8],
\]

\[
E \in [3..7], N \in [4..8],
\]

\[
E \in [4..7], N \in [4..8],
\]        

\[
E \in [4..7], N \in [5..8],
\]        

\NI 
upon which the reduction process stabilizes. At this stage the
solving rules for disequalities are not applicable either.

So using the reduction rules we reduced the original ranges to (\ref{eq:send}).
The derivation, without counting the initial applications of the transformation
rules, consists of 24 steps.

Using the {\em SUBSTITUTION\/} rule of Subsection \ref{subsec:exapr}
and obvious transformation rules that deal with bringing 
the equality constraints
to the form (\ref{eq:equality}), the original equality
constraint gets reduced to
\[
90 \cdot N + Y - (91 \cdot E + D + 10 \cdot R) = 0.
\]
Moreover, ten simple disequality constraints between the variables
$E, N, D, R$ and $Y$ are still present.

Further progress can now be obtained only by employing a splitting
rule.  The behaviour of the \eclipse{} finite domain solver is
modelled by the {\em ENUMERATION\/} rule of Subsection
\ref{subsec:exapr}. It can be shown, by mimicking the \eclipse{}
execution, that, when the rules here presented are augmented by this
rule, there exists a successful derivation for the original CSP representing the
{\em SEND + MORE = MONEY} puzzle.

\subsection{Discussion}

This concludes our presentation of the proof rules that can be used
to build a constraint solver for linear constraints over interval and finite
domains.
Such proof rules are present in one form or another within each constraint
programming system that supports linear constraints over  interval and finite
domains.

It is worthwhile to mention that the reduction rules {\em LINEAR
  INEQUALITY \ref{eq:lin}} and {\em LINEAR EQUALITY }
are simple modifications of the reduction rule introduced in
\cite[page 306]{davis87} that dealt with closed intervals of reals.
Also, as pointed out to us by Lex Schrijver, these rules are instances of
the cutting-plane proof technique used in linear programming
(see e.g. \cite[Section 6.7]{CCPS98}).

\section{Building a Constraint Solver: an Example}
\label{sec:cs}

We now show how one can use our approach to define specific constraint
solvers.  By means of example consider the constraint {\em exactly(x, l,
  z)} introduced in 
\cite{vanhentenryck-constraint-using}. It
states that for a list $l$ of variables ranging over some
fixed domain exactly $x$ of its elements equal $z$.  We assume that
this primitive can be used with $x$ a variable ranging over a subset of natural
numbers and with $z$ a variable with an unspecified domain.
(This is not a restriction since we can augment
the rules below with some natural variable introduction rules.)
The {\em exactly(x,l,z)\/} primitive is useful for dealing with
scheduling problems. 

Below we assume that $m >0$, $i \in [1..m]$ and that
${\cal DE}$ stands for a sequence of domain expressions
involving the relevant variables.
Further, given a set $D$ of natural numbers, we define
\[
D - 1 := \C{d-1 \mid d \in D, d > 0}.
\]

The behaviour of the {\em exactly(x,l,z)\/} primitive is described by
means of the following four rules:
\begin{list}%
{}{\usecounter{rulecnt}}
\centering
\item {\em EXACTLY  \ref{eq:exa04}}
  \label{eq:exa04}
\[
\frac
{\p{{\em exactly(x, [y_1, \LL, y_m], z)}}{x \in D_x, {\cal DE}}}
{\p{{\em exactly(x, [y_1, \LL, y_m], z)}}{x \in D_x - \C{d \mid d > m}, {\cal DE}}}
\]
\end{list}

\begin{list}%
{}{\usecounter{rulecnt}}
\setcounter{rulecnt}{1}
\centering
\item {\em EXACTLY  \ref{eq:exa1}}
  \label{eq:exa1}
\[
\frac
{\p{{\em exactly(x, [y_1, \LL, y_m], z)}}{y_i \in D_i, z \in D_z, {\cal DE}}}
{\p{{\em exactly(x, [y_1, \LL, y_{i-1}, y_{i+1}, \LL, y_m], z)}}
{y_i \in D_i, z \in D_z,  {\cal DE}}}
\]
\end{list}
where $D_i \cap D_z = \ES$ and $m >1$,

\begin{center}
{\em EXACTLY 3} \\
\[
\frac
{\p{{\em exactly(x, [y_1, \LL, y_m], z)}}{x \in D_x, y_i = a, z = a, {\cal DE}}}
{\p{{\em exactly(u, [y_1, \LL, y_{i-1}, y_{i+1}, \LL, y_m], z)}, u = x-1}
{u \in D_x - 1, x \in D_x, y_i = a, z = a, {\cal DE}}}
\]
\end{center}

\begin{list}%
{}{\usecounter{rulecnt}}
\setcounter{rulecnt}{3}
\centering
\item {\em EXACTLY  \ref{eq:exa4}}
  \label{eq:exa4}
\[
\frac
{\p{{\em exactly(x, [y_1, \LL, y_m], z)}}{x = 0, {\cal DE}}}
{\p{y_1 \neq z, \LL, y_m \neq z}{x = 0, {\cal DE}}}
\]
\end{list}

It is straightforward to see that these four rules are equivalence preserving.
These rules, when augmented with a modification of the 
{\em LINEAR EQUALITY \/} rule for finite domains,
{\em DISEQUALITY\/} {\em \ref{eq:diseq1}, \ref{eq:diseq}\/}
and {\em  \ref{rule:diseq4}\/} rules and the {\em ENUMERATION\/} rule, 
form a stand alone constraint solver.

Such a solver can be directly used for example to solve Latin square
puzzles.  (Recall that a Latin square of order $n$ is defined to
be an $n \times n$ array made out of the integers $1, 2, \LL, n$ with
the property that each of the $n$ symbols occurs exactly once in each
row and exactly once in each column of the array.)  

More importantly, the {\em exactly(x,l,z)\/} primitive has been used
to specify certain type of scheduling problems, such as the car
sequencing problem (see \cite{vanhentenryck-constraint-using}).
Also, it can be used in turn to define other useful constraint
primitives, such as the {\em atmost(x,l,z)\/} primitive of
\cite{vanhentenryck-constraint-using} that states that for a
list $l$ of variables ranging over some fixed domain atmost $x$ of its
elements equal $z$.  To this end it suffices to adopt the following
rule

\[
\frac
{\p{{\em atmost(x, l, z)}}{x \in D_{x}, {\cal DE}}}
{\p{{\em exactly(y, l, z)}, y \leq x}{y \in D_{x}, x \in D_{x}, {\cal DE}}}
\]
and add an instance of the {\em LINEAR INEQUALITY \ref{eq:lin2}} rule
that deals with simple inequalities of the form $y \leq x$.

\section{Conclusions}

\subsection{Summing up}

We presented here a proof theoretic framework that allows us to model
computing using constraints. 

In general, constraint programming consists of a generation of
constraints and of solving them. Both phases can be intertwined. In
our presentation we only concentrated on the latter aspect of
constraint programming. To complete the picture the framework here
presented should be combined with a specific ``host'' programming
language from which the constraints can be generated. To model computing
in such an amalgamated language the proof rules should be combined
with transitions dealing with the program state.  In such a constraint
programing language one can distinguish two computation steps.

\begin{itemize}

\item If a ``conventional'' programming statement (such as a procedure
call or a built-in in the logic programming framework,
or an assignment or a {\tt WHILE} loop in the imperative programming 
framework) is encountered, a usual transition is performed and the program
state is modified accordingly.

\item If a constraint is encountered, it is added to the current set of
constraints (``constraint store''). This addition is followed by a
repeated  application of the proof rules to the constraint store. The
order of application of these rules is determined by some built-in
scheduler (see the next subsection).
The terminating condition depends on specific applications. 

\end{itemize}

Further, in such an amalgamated language the interaction between the
constraint store and the program state should be properly taken care of.

In the case of linear constraints on finite domains the deterministic 
proof rules are
repeatedly applied  until all constraints are solved or 
a CSP is generated that is closed under the applications of these rules.
In the case of algebraic constraints on real intervals
the proof rules are  repeatedly applied until all constraints are solved
or all intervals are smaller than some fixed in advance $\epsilon$.

In some constraint programming languages or in the case of some
constraint solvers the splitting rules are not scheduled. Instead,
their application is  explicitly triggered by some programming
construct or facility present in the language.

Let us compare now in more detail our approach to that of
\cite{Cas98} and \cite{fruhwirth-constraint-95}.  In
\cite{Cas98} the proof rules are represented as rewrite rules in
the programming language {\sf ELAN} (for the most recent reference see 
\cite{BKK98}). {\sf ELAN}
allows one to define specific strategies that can be used to schedule
these rewrite rules. We defer discussion of this aspect to the
next subsection.

In constraint handling rules ({\sf CHR}s) of
\cite{fruhwirth-constraint-95} the rules manipulate constraints only,
so the domains need to be encoded as unary constraints. This leads to
a different than ours classification of rules according to which
``propagation'' means addition of redundant constraints.  Further,
these rules are not supposed to be used ``stand alone'' but rather to
augment constraint logic programming that provides already a support
for the ``don't know'' nondeterminism. So no splitting rules are
available. 

So the implementation of constraint solvers defined by the {\sf ELAN}
rules and by {\sf CHR}s is automatically provided by the interpreter,
respectively compiler of the language. In contrast,  our approach is
not geared towards direct implementability even though an
implementation specified by such rules is pretty obvious. This allows
us to be more abstract and permits us to express and analyze
specific, domain dependent, constraint solvers in a simple way.

For example we can readily define proof rules that involve some
auxiliary computations, such as the {\em LINEAR INEQUALITY\/}
reduction rule of Subsection \ref{subsec:equ} and study formally
properties of such rules (see Appendix).  Such an analysis would be
difficult to achieve if we had to reason about an encoding of these
rules in a specific programming formalism.

\subsection{Control}
\label{subsec:control}

One of the issues conspicuously absent in our considerations is that
of control. To draw the analogy with the ``Algorithm = Logic
+ Control'' slogan of \cite{Kow79a}, 
what we defined here is only the ``Logic'' part of
constraint programming. The picture is completed once we can
adequately deal with the ``Control'' part.

In the proof theoretic framework here presented the issue of control
enters the picture at three places. First, one needs to schedule the
introduced proof rules. Second, one should be able to define which
rules are to be scheduled. Finally, one needs some search strategy to
traverse the generated proof trees in search for a successful
derivation.

The scheduling of the proof rules could be done using a built-in
strategy that combines scheduling of the deterministic rules by means
of a generic chaotic iteration algorithm of \cite{Apt97b} with
the requirement that the applications of the splitting rules are
delayed as much as possible.  We noted in \cite{Apt97b} that
several constraint propagation algorithms employ in fact such a
generic algorithm. Further, delay of the applications of the splitting
rules prevents unnecessary creation of alternative branches and is a
well-known and widely used heuristic. So such a ``hard-wired''
strategy seems perfectly reasonable.

In contrast, we think that both the selection of specific proof rules
and the search strategies for traversing the proof trees should be
programmable. What we need here is a programming notation that could allow us
to define most common search strategies in a simple way.
One possibility would be to use {\sf ELAN} that,
as already mentioned, allows one to define various search strategies. 
\cite{Cas98} showed how several of them, 
such as forward checking and 
various forms of look ahead, can be implemented in {\sf ELAN}.
We believe that more work is needed to see whether other strategies
such as backjumping can be expressed in {\sf ELAN}, as well.

Another work that should be mentioned here is \cite{Sch97} where
it is shown how the concept of so-called computation spaces can be
used to program in a simple way various search strategies in the
programming language Oz (see \cite{Smo95}).

\section*{Acknowledgements}
We would like to thank Eric Monfroy for numerous discussions on the
subject of this paper and Carlos Castro, Martin Henz,  Lex Schrijver,
Gerhard Wetzel and an anonymous referee for providing
useful comments on a draft of this paper.  



\section*{Appendix: a Characterization of the {\em LINEAR EQUALITY\/} Rule}

The rules we introduced in Section \ref{sec:lin}
seem somewhat arbitrary. After all, using them we cannot even 
solve the CSP
$\p{x + y = 10, x - y = 0}{x \in [0..10], y \in [0..10]}$ as
no application of the {\em LINEAR EQUALITY\/} rule
of Subsection \ref{subsec:equ} to this CSP is relevant.

The point is that the domain reduction rules like the {\em LINEAR
EQUALITY\/} rule are in some sense ``orthogonal'' to the rules
that deal with algebraic manipulations (that can be expressed as
transformation rules): their aim is to reduce the domains and not to
transform constraints. So it makes sense to clarify what
these rules actually achieve.  This is the aim of this section.  
By means of example, we concentrate here on the {\em LINEAR EQUALITY\/} rule.

To analyze it it will be useful to
consider first its counterpart that deals with the intervals of reals.  This
rule is obtained from the {\em LINEAR EQUALITY} rule by
deleting from the definitions of $l'_j$ and
$h'_j$ the occurrences of the functions $\ceiling{\:}$ and
$\floor{\:}$. 
So this rule deals with an equality constraint in the form
\[
\sum_{i \in {\em POS}} a_i x_i -  \sum_{i \in {\em NEG}} a_i x_i = b
\]
where
\begin{itemize}
\item $a_i$ is a positive integer for $i \in {\em POS} \cup {\em NEG}$,

\item $x_i$ and $x_j$ are different variables for $i \neq j$ and $i,j
  \in {\em POS} \cup {\em NEG}$,

\item $b$ is an integer,
\end{itemize}
and intervals over reals. 

In what follows for two reals $r_1$ and $r_2$ we denote by 
$[r_1, r_2]$ the closed interval of real line bounded by $r_1$ and $r_2$.
So $\pi \in [3,4]$ while $[3..4] = \C{3,4}$.

The rule in question has the following form:

\begin{list}%
{}{\usecounter{rulecnt}}
\item {\em ${\cal R}$-LINEAR EQUALITY }
\label{eq:r-equ}
\centering
\item 
\[
\frac
{\p{\sum_{i \in {\em POS}} a_i x_i -  \sum_{i \in {\em NEG}} a_i x_i = b}{x_1 \in [l_1 .. h_1],  \LL,  x_{n} \in [l_n..h_n]}}
{\p{\sum_{i \in {\em POS}} a_i x_i -  \sum_{i \in {\em NEG}} a_i x_i = b}{x_1 \in [l^{r}_1 .. h^{r}_1],  \LL,  x_{n} \in [l^{r}_n..h^{r}_n]}}
\]
\end{list}
where for $j \in {\em POS}$
\[
l^{r}_j := max(l_j, \gamma_j), \ h^{r}_j := min(h_j, \alpha_j)
\]
and for $j \in {\em NEG}$
\[
l^{r}_j := max(l_j, \beta_j),  \ h^{r}_j := min(h_j, \delta_j).
\]
Recall that
$\alpha_j, \beta_j, \gamma_j$ and $\delta_j$ are defined in
Subsection \ref{subsec:equ}. In particular, recall that
\[
\alpha_j = \frac{b - \sum_{i \in {\em POS} - \{j\}} a_i l_i +  \sum_{i \in {\em NEG}} a_i h_i}{a_j}
\]
and
\[
\gamma_j := \frac{b - \sum_{i \in {\em POS} - \{j\}} a_i h_i +  \sum_{i \in {\em NEG}} a_i l_i}{a_j}
\]

 It is
straightforward to see that the ${\cal R}$-{\em LINEAR EQUALITY\/} rule, when interpreted over the
intervals of reals, is also equivalence preserving.

To characterize the ${\cal R}$-{\em LINEAR EQUALITY\/} rule 
we use the following notion introduced in \cite{MM88}.
The original definition for binary constraints is due to
\cite{mackworth-consistency}.

\begin{definition} \mbox{} \\[-6mm]

\begin{itemize}

\item A constraint $C$ is called {\em arc-consistent\/}
if for every variable of it each value in its domain participates
in a solution to $C$.

\item A CSP is called {\em arc-consistent\/} if every constraint of it is.
\HB
\end{itemize}
\end{definition}

We now prove the following result.

\begin{theorem} \label{thm:r-lin}
\mbox{} \\
\vspace{-6mm}
  
  \begin{enumerate} \smallromani
  \item 
The conclusion of the ${\cal R}$-{LINEAR EQUALITY} 
rule is either failed or arc consistent.  

\item In the case of a single linear equality constraint the ${\cal
    R}$-{LINEAR EQUALITY} rule is idempotent, that is a CSP is closed
  under the application of this rule after one iteration.

\item In the case of more than one linear equality constraint
the ${\cal R}$-{LINEAR EQUALITY} rule can yield an infinite derivation.

\end{enumerate}
\end{theorem}
\Proof
(i) Assume that the conclusion of the
${\cal R}$-{\em LINEAR EQUALITY\/} rule is not failed.
Fix $j \in {\em POS}$. 
We have both
\[
\sum_{i \in {\em POS} - \C{j}} a_i l_i + a_j \alpha_j  - \sum_{i \in {\em NEG}} a_i h_i = b
\]
and
\[
\sum_{i \in {\em POS} - \C{j}} a_i h_i + a_j \gamma_j  - \sum_{i \in {\em NEG}} a_i l_i = b.  
\]

Hence for any $\alpha$
\[
\alpha ( \sum_{i \in {\em POS} - \C{j}} a_i (l_i - h_i) + a_j (\alpha_j - \gamma_j) - \sum_{i \in {\em NEG}} a_i (h_i - l_i)) = 0,
\]
so for any $\alpha$
\begin{equation}
  \label{eq:lin1}
\sum_{i \in {\em POS} - \C{j}} a_i (h_i + \alpha (l_i - h_i)) + a_j (\gamma_j + \alpha(\alpha_j - \gamma_j)) - \sum_{i \in {\em NEG}} a_i (l_i + \alpha (h_i - l_i)) = b.
\end{equation}

By the definition of  $l^{r}_j$ and $h^{r}_j$ 
we have
$[l^{r}_j, h^{r}_j] \sse [\gamma_j, \alpha_j]$. By
the initial assumption the interval $[l^{r}_j, h^{r}_j]$ is non-empty. 

Take now some $d \in [l^{r}_j, h^{r}_j]$.
Next, take $\alpha$ such that $\gamma_j + \alpha(\alpha_j - \gamma_j) = d$,
that is
\[
\alpha := \frac{d - \gamma_j}{\alpha_j - \gamma_j}
\]
and choose the solution to 
$\sum_{i \in {\em POS}} a_i x_i -  \sum_{i \in {\em NEG}} a_i x_i = b$
determined by $\alpha$ and (\ref{eq:lin1}).

By the choice of $d$ 
and the fact that $[l^{r}_j, h^{r}_j]$ is non-empty,
$\alpha$ is well-defined and $0 \leq \alpha \leq 1$.
Hence this solution to 
$
\sum_{i \in {\em POS}} a_i x_i -  \sum_{i \in {\em NEG}} a_i x_i = b
$
lies in the ranges
$[l_i .. h_i]$ with $i \in [1..n]$.
But the ${\cal R}$-{\em LINEAR
  EQUALITY\/} rule is equivalence preserving, so this solution
also lies in the ranges
$[l^{r}_i .. h^{r}_i]$ with $i \in [1..n]$.
This proves the claim.
\III

\NI
(ii) Straightforward by (i) as the domains of an arc consistent CSP
cannot be reduced without losing equivalence.
\III

\NI
(iii) 
The following example is due to \cite[page 304]{davis87}.  Take
$\p{x = y, x = 2y}{x \in [0,100], y \in [0,100]}$ and consider a
derivation in which the constraints are selected in an alternating fashion.
This derivation is easily seen to be infinite.
\HB
\VV

Let us mention here that property (i) generalizes a
corresponding result stated in \cite[page 326]{davis87} 
for the more limited case of so-called unit coefficient constraints.

In contrast, the {\em LINEAR EQUALITY} rule behaves differently.
First, it is not idempotent even in the case of a single linear
equality --- it just suffices to see the example at the end of
Subsection \ref{subsec:equ}.  Second, in the case of several linear
equality constraints its repeated use always terminates (due to the
fact that the domains are finite) and yields a CSP that is closed
under the applications of this rule (by the fact that it is
equivalence preserving).

Further, the example at the end of Subsection \ref{subsec:equ} also shows
that a CSP closed under the applications of this rule does
not need to be arc consistent. So we need another notion
to characterize CSP's closed under the applications of this rule.

First, we introduce the following terminology.

\begin{definition} \mbox{} \\[-6mm]
\begin{itemize}

\item  By an {\em ICSP\/} we mean a CSP the
  domains of which are intervals of reals or integers.

\item A constraint $C$ on a non-empty sequence of variables, the
  domains of which are intervals of reals or integers, is called {\em
    bound consistent\/} if for every variable of it each of its two
  bounds participates in a solution to $C$.

\item An ICSP is called {\em bound consistent\/} if every constraint
  of it is.  
\HB
\end{itemize}
\end{definition}

This notion is motivated by a similar concept introduced in 
\cite{Lho93} in the case of constraints on reals.
Note that if a constraint with interval domains is  bound consistent,
then both the constraint and each of its intervals is non-empty.

Denote now by LINEQ an  ICSP all constraints of which are linear
equalities of the form (\ref{eq:equality}) of Subsection
\ref{subsec:equ} and discussed above.

\begin{definition} 

Consider a LINEQ $\phi$.
Let $\phi^{r}$ denote the CSP obtained from $\phi$ by replacing each integer
domain $[l..h]$ by the corresponding interval $[l,h]$ of reals.
We say that  $\phi$ is {\em interval consistent\/} if  $\phi^{r}$
is bound consistent.
\HB
\end{definition}

So $\phi$ is interval consistent if for every constraint $C$ of it 
the following holds:
for every variable of $C$ each of its two bounds participates in a
solution to $\phi^{r}$.

For example, the CSP 
$\p{3x - 5y = 4}{x \in [3..9], y \in [1..4]}$
of Subsection \ref{subsec:equ}
is bound consistent as both $x =3, y = 1$ and $x = 8, y = 4$
are solutions of $3x - 5y = 4$.

In contrast, the CSP $\phi := \p{2x + 2y - 2z = 1}{x \in [0..1], y \in
  [0..1], z \in [0..1]}$ is clearly
not bound consistent but it is interval
consistent. Indeed, the equation $2x + 2y - 2z = 1$ has three solutions
in the unit cube formed by the real unit intervals for $x,y$ and $z$:
(0,0.5,0), (1,0,0.5) and (0.5,1,1). So each bounds participates in a
solution to 
$\phi^r := \p{2x + 2y - 2z = 1}{x \in [0,1], y \in [0,1],  z \in [0,1]}$.

The following result now characterizes the outcome of a repeated
application of the {\em LINEAR EQUALITY\/} rule.

\begin{theorem} \label{thm:lin}

Consider a LINEQ $\phi$ that is closed under the applications of
the LINEAR EQUALITY rule. Then
$\phi$ is either failed or interval consistent.  
\end{theorem}

\Proof
The intervals of reals corresponding to the integer intervals in the 
conclusion of the {\em LINEAR EQUALITY\/} rule are respectively
smaller than those in the conclusion of the ${\cal R}$-{\em LINEAR EQUALITY\/} 
rule. So the assumption implies that $\phi^r$ is closed under the
applications of the ${\cal R}$-{\em LINEAR EQUALITY\/} rule. 

Assume now that $\phi$ is not failed. Then $\phi^r$ is not failed either.
By Theorem \ref{thm:r-lin}
$\phi^r$ is arc consistent, so a fortiori bound consistent.
\HB
\VV

The example preceding the above theorem shows that interval consistency
cannot be replaced here by bound consistency. In other words, to characterize
the {\em LINEAR EQUALITY\/} rule it is needed to resort to a study
of solutions over reals.

Consider now a  LINEQ CSP $\phi$ and a
derivation that consists solely of the applications of
the {\em LINEAR EQUALITY} rule.  As noticed before it is finite. Let
$\psi$ be the final LINEQ CSP. So $\psi$ is closed under the
applications of the {\em LINEAR EQUALITY} rule.  By
Theorem \ref{thm:lin}  $\psi$ is failed or interval consistent.  

Assume now that $\phi$ is not failed.
Using the results of \cite{Apt97b} (more specifically Theorem 13 on
page 47), we can then characterize $\psi$ as the largest interval
consistent LINEX CSP that is smaller than $\phi$ and equivalent to it.
(The equivalence of $\psi$ and $\phi$ is due to the fact that the {\em
  LINEAR EQUALITY} rule is equivalent preserving.) 
This paper also
shows how the applications of the {\em LINEAR EQUALITY } rule can be
scheduled in a meaningful way by means of a generic chaotic iteration
algorithm.  The details should be clear to any reader of this paper
but a detailed exposition here would take us too far afield.

Similar characterizations results can be envisaged for other
proof systems characterizing specific type of constraints. Such
characterizations have been considered in the literature albeit not in
a proof theoretical framework.

Let us cite just two examples. In \cite{Reg94} the constraint
primitive ${\em alldistinct([x_1, \LL, x_n])}$ that states that the
listed variables are all distinct was characterized using the
arc consistency notion. 

Next, in
\cite{BGC97} the constraint primitive 
${\em sort([x_1, \LL,   x_n], [y_1, \LL, y_n])}$
was characterized using the
bound consistency notion.  This constraint states that the
second list is the sorted version of the first list; the variables
are assumed to range over intervals.  

In each case an algorithm was provided that reduces the domains of the
constraint under consideration so that the appropriate local
consistency notion is satisfied.  On a sufficiently abstract level
these algorithms can be explained by means of reduction rules.

We believe that such results allow us both to clarify and to characterize
existing constraint solvers and to look for new ones. In this respect
an  interesting question is in what sense the rules of Section
\ref{sec:cs} characterize the {\em exactly(x,l,z)\/} primitive.

\end{document}